# Hierarchical Multi-Agent Framework for Carbon-Efficient Liquid-Cooled Data Center Clusters

**Soumyendu Sarkar**[1*†], **Avisek Naug**[1*], **Antonio Guillen**[1*], **Vineet Gundecha**[1*],
**Ricardo Luna Gutiérrez**[1], **Sahand Ghorbanpour**[1], **Sajad Mousavi**[1], **Ashwin Ramesh Babu**[1],
**Desik Rengarajan**[12], **Cullen Bash**[1]

[1]Hewlett Packard Enterprise
[2]Amazon

{soumyendu.sarkar, avisek.naug, antonio.guillen, vineet.gundecha, rluna, sahand ghorbanpour, sajad.mousavi,
ashwin.ramesh-babu, desik.rengarajan, cullen.bash}@hpe.com

## Abstract

Reducing the environmental impact of cloud computing requires efficient workload distribution across geographically dispersed Data Center Clusters (DCCs) and simultaneously optimizing liquid and air (HVAC) cooling with time shift of workloads within individual data centers (DC). This paper introduces Green-DCC, which proposes a Reinforcement Learning (RL) based hierarchical controller to optimize both workload and liquid cooling dynamically in a DCC. By incorporating factors such as weather, carbon intensity, and resource availability, Green-DCC addresses realistic constraints and interdependencies. We demonstrate how the system optimizes multiple data centers synchronously, enabling the scope of digital twins, and compare the performance of various RL approaches based on carbon emissions and sustainability metrics while also offering a framework and benchmark simulation for broader ML research in sustainability.

## Introduction

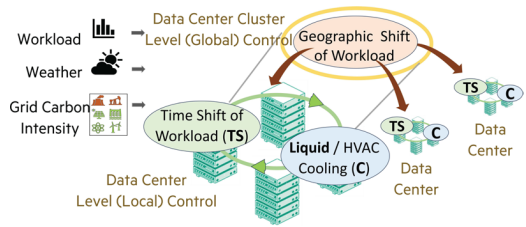

Figure 1: Spatio temporal hierarchical workload distribution.

To address these sustainability challenges with the exponential growth of AI and digital services, major cloud service providers with geographically distributed data centers are exploring innovative approaches to optimize workload distribution and energy efficiency (1; 4; 2; 8; 5).
The challenges are at two levels: the individual data center level (low level) and the data center cluster level (top level). As discussed in our previous work on holistic real-time data center control (17; 16; 18; 7), there are complex internal dependencies between multiple controls, such as time shift of workload and cooling, and dynamic external dependencies, such as workload, weather, and carbon intensity. This requires an adaptive approach like multiagent reinforcement learning (MARL) that static solutions cannot provide. While the top-level control attempts to optimally distribute the load between DCs, the low-level controller at the DC level time shifts the workload, which causes difficult synchronization and control. Note that there has been substantial progress in Reinforcement Learning across various fields, from complex control systems to training LLMs (19; 3; 12; 14; 9; 11; 13; 15; 10; 6).

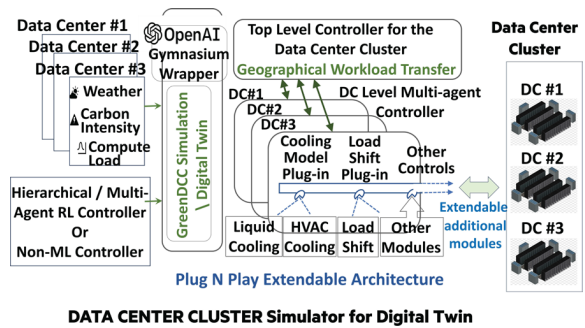

Figure 2: Green-DCC framework for Hierarchical Data Center Cluster Control leveraging geographic differences in weather, grid carbon intensity, etc. DC level modeling follows (7).

In addition, liquid cooling, with its better thermal conductivity compared to air, offers significant energy savings and a reduction in carbon footprint, particularly in GPU-enabled servers and high-performance computing. We explore how RL control can significantly increase cooling efficiency by about 20% by effectively controlling pump flow, temperature setpoint, and blade level flow control. By combining advanced cooling technologies with intelligent spatio-temporal workload distribution (Figure 1, and 3), our aim is to maximize the overall sustainability of data center operations.

---

[*]These authors contributed equally.
[†]Corresponding Author

## Proposed Method

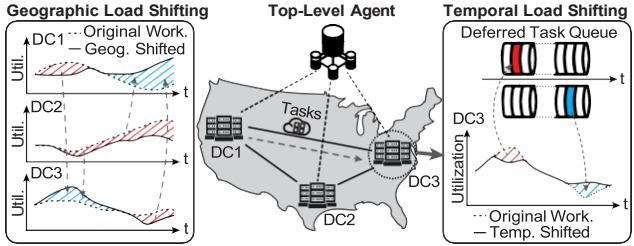

Figure 3: Left: Geographic load shifting between data centers (DC1, DC2, DC3). Right: Temporal load shifting within a data center.

The architecture employs a hierarchical RL framework to optimize workload distribution across geographically dispersed data centers by a global agent, followed by temporal rescheduling of the loads at each DC by a dedicated agent. Tiered decisions at the geographic and local data center levels are based on carbon intensity and the workload assigned to each data center. Figure 1 illustrates the geographical load shifting strategy, where tasks are dynamically moved between data centers to optimize resource utilization and reduce environmental impact. Lower-level agents handle localized operations within each data center, focusing on temporal workload scheduling. Noncritical tasks are deferred using a Deferred Task Queue (DTQ) for processing during low grid carbon intensity time periods. Figure 3 shows this combined spatiotemporal load shifting strategy to optimize data center operations' sustainability and efficiency at both global and local levels. We also introduce a dynamic cost model for workload transfers between data centers that account for geographical distance, transfer capacity, and resource constraints, balancing carbon efficiency with effective resource utilization.

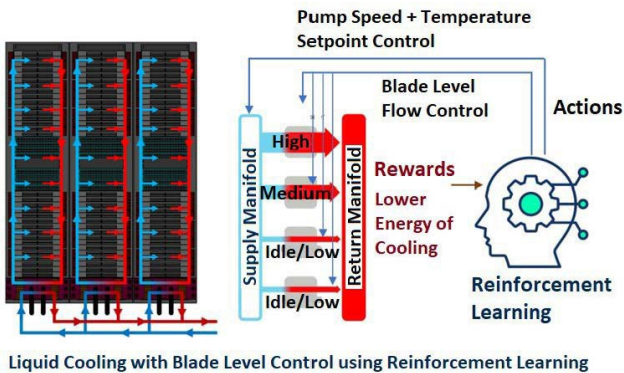

Figure 4: Blade Level RL control for Liquid Cooling

AI workload with GPUs, high performance computing, and high density servers, along with the sustainability mandate, have triggered growing adoption of high efficiency liquid cooled (LC) data centers. However, effective control of LC is lacking, with most industry implementations using open-loop control. Our work proposes RL agents to dynamically control liquid flow and temperature setpoints. Additionally, this RL agent controls the liquid flow all the way to the blade level for the emerging LC data centers (Figure 4).

Overall, the combination of a top-level geographical load shifting agent followed by a temporal load shifting and cooling optimizer agent at each data center in a hierarchical framework allows us to perform holistic optimization for data center clusters leveraging geographic differences in external parameters like weather and power grid carbon intensity. We provide a simulation framework to enable other ML researchers as shown in Figure 2. The code, licenses, and instructions for using Green-DCC can be found on GitHub[1]. Documentation can be found here[2].

| Config | Agent | | | |
|---|---|---|---|---|
| | Baseline | Top RL | Top+Low RL | HRL |
| $CO_2$ | 3845 ± 3 | 3515 ± 3 | 3512 ± 3 | 3435 ± 5 |

Table 1: Evaluation with load shift only with Hierarchical RL (HRL) showing best results.

| Config. | Hierarchical RL Algorithm | | |
|---|---|---|---|
| | PPO | A2C | APPO |
| $CO_2$ | 3435 ± 5 | 3601 ± 3 | 3608 ± 3 |

Table 2: Tones of CO2 produced by different RL algorithms. ± shows the standard deviation over the 10 seeds.

## Results

We evaluated a data center cluster with locations in New York, California, and Georgia, each with unique grid carbon intensity, weather, and workload utilization patterns. As shown in Table 1, three variations of the RL algorithm were tested: (1) Top-Level RL, which trains a top-level agent for geographic workload distribution without lower-level temporal load shifting; (2) Top + Low-Level RL, combining geographic and temporal distribution with a pre-trained low-level agent; and (3) Hierarchical RL (HRL), where all agents are trained simultaneously.
PPO (19) yielded the best results in all cases. Only load shifts were used for comparison. Although hierarchical RL is the most complex to train, it showed the best performance with PPO, minimizing the most CO2 emissions. The results are presented in Tables 1 and 2.

## Conclusion

Green-DCC represents a significant step forward in comprehensive spatio-temporal workload optimization in data center clusters using hierarchical reinforcement learning. By providing a realistic, comprehensive benchmark environment that captures the complexities of geographical and temporal workload shifting, we enable researchers and practitioners to develop innovative solutions that can dramatically reduce the carbon footprint of data center clusters.

---

[1]GitHub repository: github.com/HewlettPackard/green-dcc.
[2]Documentation: hewlettpackard.github.io/green-dcc.

## Acknowledgments

We would like to thank Paolo Faraboschi for sharing his expertise in machine learning and practical implementation approaches and Torsten Wilde for his feedback on energy optimization and sustainability.

Furthermore, we extend our gratitude to Wes Brewer, Feiyi Wang, Vineet Kumar, Scott Greenwood, Matthias Maiterth, and Terry Jones of Oak Ridge National Laboratory for their feedback and leadership within the ExaDigiT consortium, which helped refine our solution.